\documentclass{article}
\usepackage[T1]{fontenc}
\usepackage{iclr2027_conference,times}

\usepackage{amsmath,amsfonts,bm}

\def\eqref#1{equation~\ref{#1}}

\def\1{\bm{1}}

\def\vc{{\bm{c}}}

\def\vv{{\bm{v}}}

\def\vz{{\bm{z}}}

\DeclareMathAlphabet{\mathsfit}{\encodingdefault}{\sfdefault}{m}{sl}
\SetMathAlphabet{\mathsfit}{bold}{\encodingdefault}{\sfdefault}{bx}{n}

\usepackage{amsmath,amssymb}
\usepackage{booktabs}
\usepackage{multirow}
\usepackage{longtable}
\usepackage{graphicx}
\usepackage{microtype}
\usepackage{placeins}
\usepackage{xcolor}
\usepackage{tikz}
\usetikzlibrary{arrows.meta}
\usepackage{hyperref}
\usepackage{url}
\hypersetup{
  colorlinks=true,
  allcolors=blue,
  pdftitle={Triangular Resampling for Long-Horizon Motion Generation},
  pdfauthor={Kunhang Li, Yiyi Cai, Xiangyue Zhang, Fangyuan Tu, Yuhan Wu, Zhixiang Wang, Kaipeng Zhang, Haiyang Liu}
}

\title{Triangular Resampling for\\Long-Horizon Motion Generation}

\author{%
  Kunhang Li\textsuperscript{1,2,}\thanks{Work done during internship at Alaya Lab.}\quad
  Yiyi Cai\textsuperscript{2}\quad
  Xiangyue Zhang\textsuperscript{1,2}\quad
  Fangyuan Tu\textsuperscript{2,3}\\
  \bfseries Yuhan Wu\textsuperscript{1}\quad
  Zhixiang Wang\textsuperscript{2}\quad
  Kaipeng Zhang\textsuperscript{2,}\thanks{Corresponding authors.}\quad
  Haiyang Liu\textsuperscript{1,2,}\footnotemark[3]\\[4pt]
  \normalfont\textsuperscript{1}The University of Tokyo\quad
  \textsuperscript{2}Alaya Lab\\
  \normalfont\textsuperscript{3}Japan Advanced Institute of Science and Technology
}

\iclrfinalcopy

\newcommand{\method}{Triangular Resampling}
\newcommand{\shortmethod}{TR}

\newcommand{\metricci}[2]{\ensuremath{#1^{\scriptscriptstyle\pm #2}}}

\newcommand{\valpha}{\boldsymbol{\alpha}}
\newcommand{\vepsilon}{\boldsymbol{\epsilon}}

\begin{document}

\setcounter{footnote}{1}
\maketitle
\lhead{Preprint}
\vspace{-24pt}

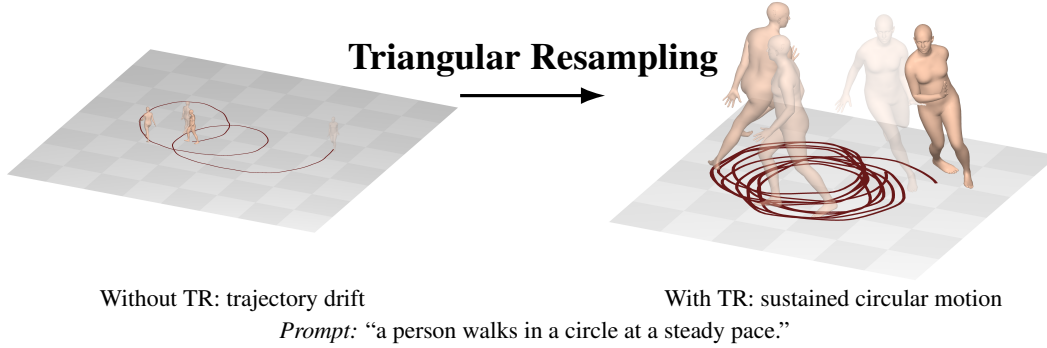
\begin{figure}[h!]
  \centering
  \begin{tikzpicture}
    \node[inner sep=0,anchor=west] (baseline) at (0,0) {\includegraphics[width=0.43\linewidth]{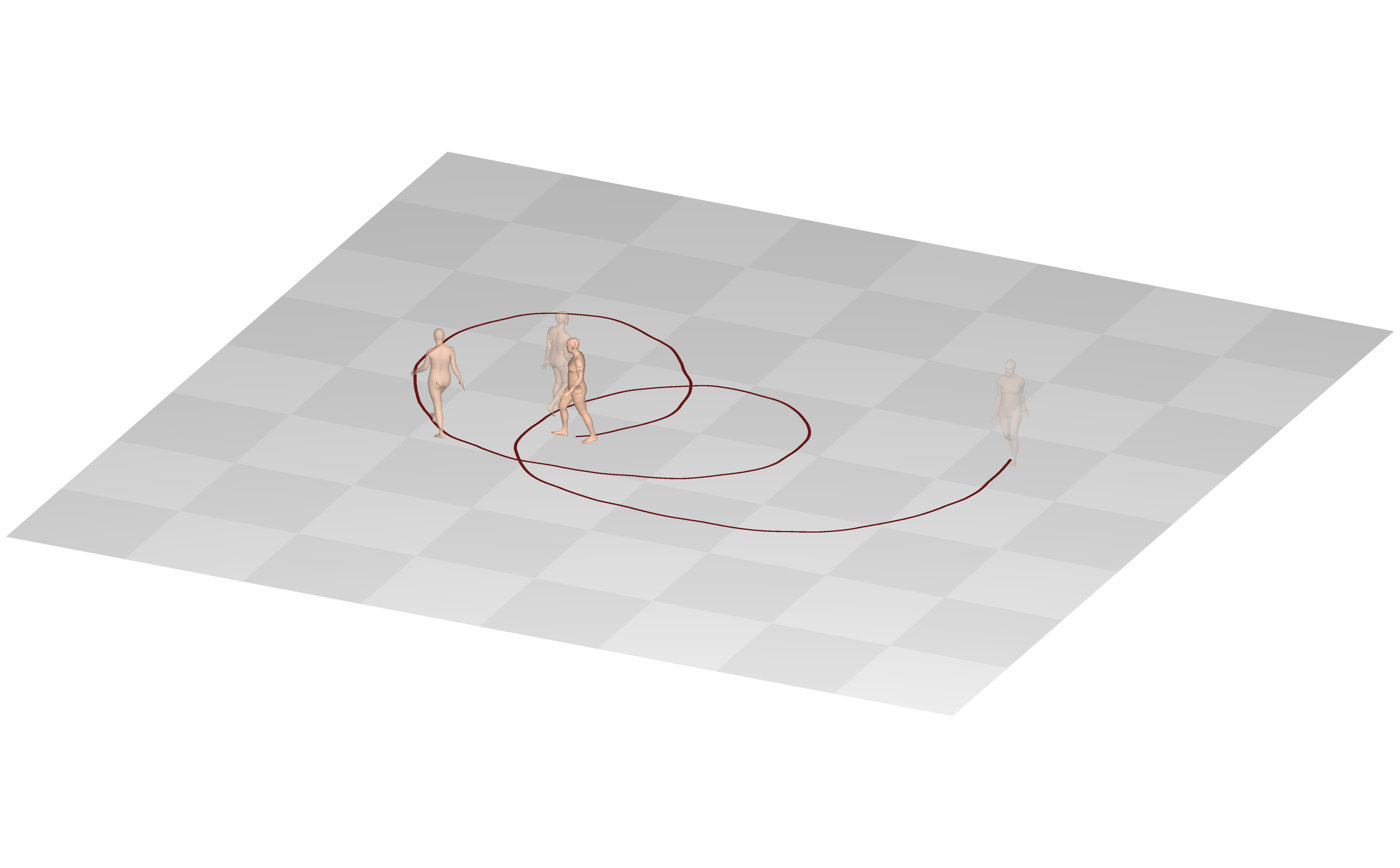}};
    \node[inner sep=0,anchor=east] (ours) at (\linewidth,0) {\includegraphics[width=0.43\linewidth]{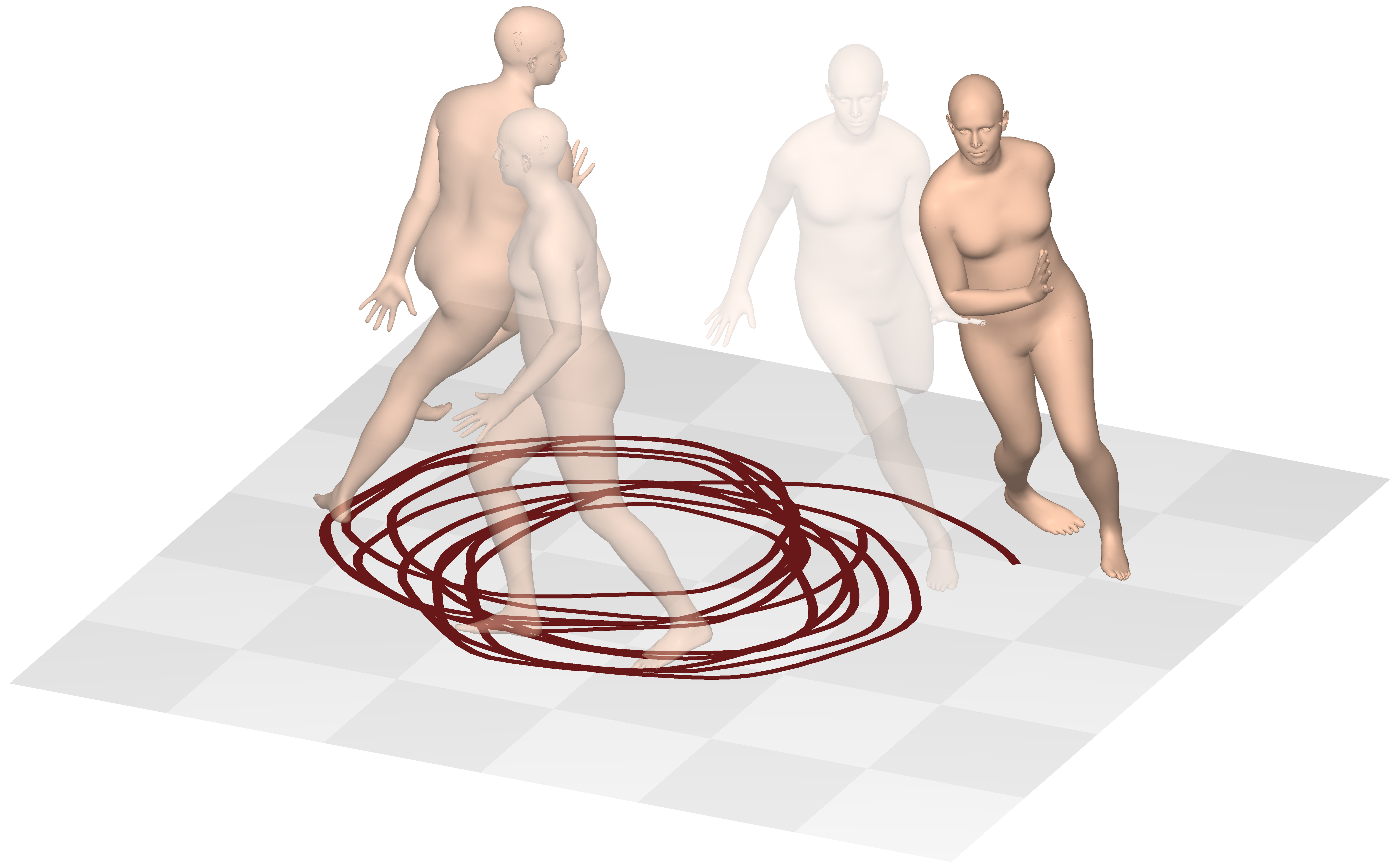}};
    \node[anchor=north,align=center,font=\small,inner sep=0] at ([yshift=-1mm]baseline.south) {Without TR: trajectory drift};
    \node[anchor=north,align=center,font=\small,inner sep=0] at ([yshift=-1mm]ours.south) {With TR: sustained circular motion};
    \draw[-{Latex[length=3.3mm,width=2.25mm]},line width=1.2pt]
      ([yshift=6mm]baseline.east) -- node[above=1.2mm,font=\fontsize{14}{16}\selectfont\bfseries,fill=white,inner xsep=1.5pt] {\method{}} ([yshift=6mm]ours.west);
  \end{tikzpicture}
  \vspace{-1mm}
  {\small\textit{Prompt:} ``a person walks in a circle at a steady pace.''\par}
  \caption{\textbf{\method{} (TR) improves long-horizon consistency.} Given the prompt ``a person walks in a circle at a steady pace,'' the original model fails to sustain circular motion over time (left), while the same model fine-tuned with \shortmethod{} better maintains a circular trajectory (right). Red curves show the ground-plane root trajectories; earlier poses are more transparent and later poses are more opaque.}
  \label{fig:teaser}
\end{figure}

\begin{abstract}
We introduce \method{} (\shortmethod{}), a post-training method for mitigating long-horizon error accumulation in motion diffusion models. TR builds on FloodDiffusion~\citep{cai2025flooddiffusion}, which generates motion using a triangular denoising schedule. During training, the current model observes short, ground-truth-derived motion windows, whereas long-horizon inference repeatedly conditions on its own predictions, allowing errors to accumulate. A common remedy is to expose the model to its own rollouts during training~\citep{zhao2025dart,xiao2025motionstreamer}. However, replacing only completed motion history does not capture the evolving, partially denoised states within a triangular denoising window.
To address this mismatch, TR introduces a key design that extends rollout-based training to the entire active window, including partially denoised states. This exposes training to model-induced errors, but unrestricted rollout can also move the training states away from their paired ground-truth motion. We therefore introduce a denoising threshold to retain ground-truth anchoring while controlling the transition to model-generated rollouts. For each replayed training sample, TR draws a shared threshold and replays the multi-step triangular denoising trajectory without gradient tracking. After each update, states below the threshold are replaced with noise-matched ground truth, while states at or above it retain model predictions. The resulting latent window is then used in the standard forward pass, loss computation, and gradient update. This rollout construction supports both the original supervised objective (\shortmethod{}) and distribution matching (TR-DMD).
We evaluate 120-second motion generation on HumanML3D test prompts. TR and TR-DMD achieve state-of-the-art FID AUC within their respective non-DMD and DMD comparison groups. Specifically, supervised \shortmethod{} reduces FID AUC by 40.9\% and FID degradation slope by 55.3\% relative to matched post-training without replay.
\end{abstract}

\section{Introduction}
\label{sec:introduction}

Long-horizon text-to-motion generation~\citep{athanasiou2022teach,shafir2024prior,barquero2024flowmdm,lee2024t2lm} requires a model to sustain plausible movement and adherence to a text instruction far beyond the short sequences used for training. A central challenge is the mismatch between training on short, ground-truth-derived motion windows and repeatedly conditioning on model predictions during inference. Each prediction becomes part of the input for subsequent generation, allowing small errors to propagate and accumulate over time. Addressing this mismatch requires considering the states the model actually encounters throughout a long rollout.

\begin{figure*}[t!]
  \centering
  \includegraphics[page=1,width=\textwidth,trim=60 234 105 50,clip]{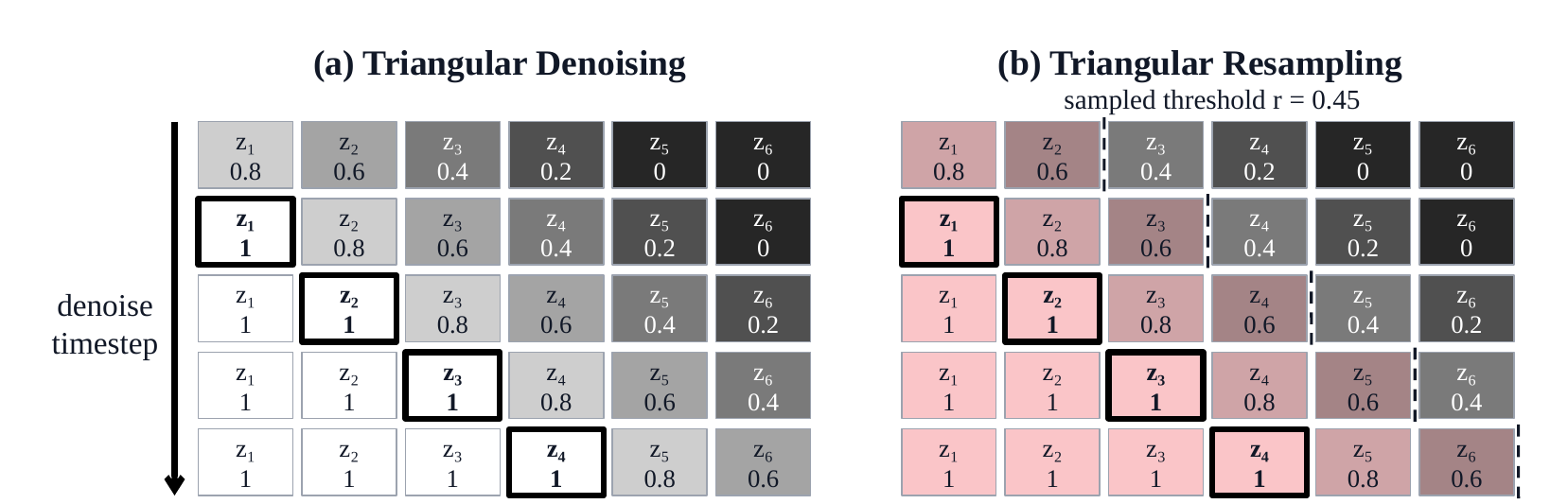}
  \caption{Comparison of triangular denoising and triangular resampling during training. The number in each cell is its denoising level $\alpha$; light cells contain less noise and dark cells more. (a) Each pair of consecutive displayed rows represents two Euler updates; the intervening update is omitted for visual clarity. Each latent undergoes ten Euler updates in total, with $\Delta\alpha=0.1$ per update and a $0.2$ offset between adjacent latents; the thick border marks the newly committed latent. (b) \shortmethod{} replays the same trajectory with one shared per-sample threshold $r$ (sampled as $0.45$ here). For latent $j$, model rollout follows $\vz^{k+1}_{j}=\vz^{k}_{j}+(\alpha^{k+1}_{j}-\alpha^{k}_{j})v_{\theta,j}(\vz^{k},\valpha^{k},\vc)$. After the update, \shortmethod{} sets $\widetilde{\vz}^{k+1}_{j}=\alpha^{k+1}_{j}\vz^{\mathrm{GT}}_j+(1-\alpha^{k+1}_{j})\vepsilon^{k+1}_{j}$ when $\alpha^{k+1}_{j}<r$ (black-to-gray GT clamp), and $\widetilde{\vz}^{k+1}_{j}=\vz^{k+1}_{j}$ when $\alpha^{k+1}_{j}\geq r$ (black-to-red model rollout), where $\vepsilon^{k+1}_{j}\sim\mathcal{N}(0,I)$. Replay repeats with the same threshold until the sampled time $t$.}
  \label{fig:overview}
\end{figure*}

We study this mismatch in FloodDiffusion~\citep{cai2025flooddiffusion}, a state-of-the-art motion diffusion model with a triangular denoising schedule (Figure~\hyperref[fig:overview]{\ref*{fig:overview}(a)}). This representative schedule for sequential generation assigns increasing noise from earlier to later latents in the active window, reflecting the intuition that near-future motion is more constrained by the committed history. During inference, the leading token is committed when it becomes clean, while the remaining tokens continue denoising as the window advances. These partially denoised states retain predictions from earlier updates and also influence subsequent ones through bidirectional attention. Standard training, however, constructs window states directly by corrupting ground-truth latents to their prescribed noise levels, without the preceding model updates. Exposing the model to its own rollouts during training is a common remedy~\citep{zhao2025dart,xiao2025motionstreamer}, but replacing only completed history leaves this mismatch unresolved in the partially denoised region.

We propose \method{} (\shortmethod{}), a post-training method that mitigates long-horizon error accumulation by addressing the train--inference mismatch in partially denoised states under triangular denoising. Its key design extends rollout-based training to partially denoised states while controlling their transition from ground-truth-derived states to model predictions (Figure~\hyperref[fig:overview]{\ref*{fig:overview}(b)}).
Unrestricted rollout can move the training states away from their paired ground-truth motion. For each replayed sample, \shortmethod{} draws one denoising threshold, shared across all tokens and Euler updates, and replays the current model's multi-step triangular trajectory without gradient tracking. After each update, states below the threshold are replaced with noise-matched ground truth, while states at or above it retain model predictions. As denoising progresses, tokens cross the shared threshold and remain under model control. In this way, \shortmethod{} preserves a ground-truth reference to limit excessive drift during replay, while exposing training to model-induced errors to address the training--inference mismatch.

The rollout construction supports two training objectives. In supervised \shortmethod{}, the replayed latent window enters the standard training update, with targets derived from the paired clean motion. Alternatively, Distribution Matching Distillation (DMD)~\citep{yin2024onestep} aligns the distribution of generated samples with that of a pretrained diffusion teacher. We combine \shortmethod{} with DMD to obtain TR-DMD, adopting the DMD update recipe of Rolling Forcing~\citep{liu2026rolling} while retaining triangular replay.

We train on HumanML3D~\citep{guo2022humanml3d} and BABEL~\citep{punnakkal2021babel}, and evaluate 120-second motion generation on HumanML3D test prompts by computing metrics in twelve non-overlapping 10-second windows. Sharing the same backbone, initialization, and 30k post-training budget, \shortmethod{} reduces FID AUC from 1.951 to 1.153 (40.9\%) and the FID linear degradation slope from 0.526 to 0.235 per minute (55.3\%) compared with the baseline without \shortmethod{} (TR-off). Both \shortmethod{} and TR-DMD achieve state-of-the-art FID AUC within their respective non-DMD and DMD comparison groups under this protocol: 1.153 for \shortmethod{} and 1.324 for TR-DMD, compared with 1.351 for Rolling Forcing in the DMD group. A complementary pairwise preference evaluation ranks \shortmethod{} highest in both motion quality and text alignment, with Bradley--Terry scores of 0.302 and 0.317, respectively. The corresponding scores are $-0.272$ and $-0.211$ for TR-off, and $-0.030$ and $-0.106$ for Rolling Forcing.

Our contributions are:

\begin{itemize}
  \item We identify a training--inference mismatch in the partially denoised states of triangular denoising and propose \method{} (\shortmethod{}) to mitigate the resulting long-horizon error accumulation. \shortmethod{} extends rollout-based training to these states, using a shared threshold to balance ground-truth anchoring and exposure to model predictions. This rollout construction supports both supervised training and distribution matching (TR-DMD).
  \item We systematically evaluate \shortmethod{} and TR-DMD for long-horizon motion generation. Both achieve state-of-the-art FID AUC within their respective non-DMD and DMD comparison groups under our evaluation protocol, with supervised \shortmethod{} improving over matched TR-off by 40.9\%. Ablations examine the effects of clamping and replay frequency, while pairwise preference evaluation further supports \shortmethod{}'s motion quality and text alignment.
\end{itemize}

\section{Related Work}
\label{sec:related}

\paragraph{Long-Horizon Motion Generation.}
TEACH~\citep{athanasiou2022teach} composes motion segments autoregressively, while DoubleTake~\citep{shafir2024prior} refines overlapping segments from a short-motion diffusion prior. FlowMDM~\citep{barquero2024flowmdm} and T2LM~\citep{lee2024t2lm} further study long-form motion synthesis through positional encoding and latent composition, respectively. FloodDiffusion~\citep{cai2025flooddiffusion} jointly denoises a bidirectional active window under a triangular schedule. Rather than proposing another composition or sampling architecture, we optimize the training-state distribution of this triangular generator for long-horizon motion.

\paragraph{Rollout Training in Motion Generation.}
DART~\citep{zhao2025dart} trains on overlapping motion primitives with a staged curriculum that progresses from ground-truth histories through mixed histories to full diffusion rollouts. MotionStreamer~\citep{xiao2025motionstreamer} uses Two-Forward training: an initial prediction pass supplies latents that replace a progressively increasing subset of history tokens before the standard training update. PRISM v1~\citep{ling2026prismv1} applies Self Forcing~\citep{huang2025selfforcing} to generated segments, decoding and re-encoding each segment as context for the next and optimizing the resulting rollouts through distribution matching. These approaches expose motion models to their own predictions through history replacement or self-conditioned rollout training. Our focus is the evolving joint state of a triangular denoising window, including both committed outputs and partially denoised tokens.

\paragraph{Rollout Training in Video Generation.}
Self Forcing~\citep{huang2025selfforcing} uses few-step autoregressive self-rollouts with truncated gradients and optimizes a video-level distribution-matching loss. Self Gradient Forcing~\citep{zhuang2026selfgradient} recomputes historical key--value representations in a parallel second pass, allowing future DMD~\citep{yin2024onestep} losses to train context encoding without differentiating through the serial rollout. Rolling Forcing~\citep{liu2026rolling} jointly denoises a rolling window at staggered noise levels, retains initial-frame attention sinks, and mixes rolling-window and Self Forcing DMD updates with equal probability. Causal Forcing~\citep{zhu2026causalforcing} uses an autoregressive teacher for ODE initialization, followed by Self Forcing-style DMD. Resampling Forcing~\citep{guo2025resampling} instead resamples noise-corrupted ground-truth frames autoregressively with the current model, then conditions training on the detached resampled histories while retaining the original ground-truth-supervised flow-matching objective. It requires neither an auxiliary teacher nor a discriminator.
Like Rolling Forcing, the FloodDiffusion backbone jointly denoises a window at staggered noise levels. Like Resampling Forcing, supervised \shortmethod{} uses detached model-generated states and retains ground-truth supervision. \shortmethod{} applies a shared denoising threshold throughout triangular replay. After each update, states below the threshold return to noise-matched ground truth, while states at or above it retain model predictions.

\section{Preliminaries: FloodDiffusion Triangular Denoising}
\label{sec:preliminaries}

Let $\vz^{\mathrm{GT}}_{1:T}$ be clean motion latents and $\vc$ the text condition. FloodDiffusion~\citep{cai2025flooddiffusion} uses a linear flow-matching path~\citep{lipman2023flowmatching} with a triangular denoising schedule. Let $\tau$ denote its continuous global denoising phase. The clean-data coefficient of latent position $j$ is
\begin{equation}
  \alpha_j(\tau)=\operatorname{clip}\!\left(\tau-\frac{j}{c},0,1\right),
  \label{eq:triangular_schedule}
\end{equation}
where $c$ controls the slope and the number of simultaneously active tokens. The clean and noisy boundaries are
\begin{equation}
  m(\tau)=\left\lceil(\tau-1)c\right\rceil,
  \qquad
  n(\tau)=\left\lceil\tau c\right\rceil,
  \label{eq:active_boundaries}
\end{equation}
so positions before $m(\tau)$ are clean, positions from $m(\tau)$ to $n(\tau)-1$ form the active denoising window, and later positions remain noise. This produces the triangular schedule in Figure~\hyperref[fig:overview]{\ref*{fig:overview}(a)}.

At inference, let $k$ denote the Euler integration step and $\tau_k$ the corresponding global denoising phase. The solver advances
\begin{equation}
  \tau_{k+1}=\tau_k+\Delta\tau,
  \qquad \Delta\tau=\frac{1}{N},
  \qquad \alpha_j^k:=\alpha_j(\tau_k),
  \label{eq:phase_update}
\end{equation}
where $N$ is the number of Euler steps per unit phase. Each step therefore advances the triangular frontier by $c\Delta\tau=c/N$ positions. Token $j$ reaches the clean endpoint at $\tau=1+j/c$, so successive commits are separated by $1/c$ in phase, or $N/c$ Euler steps when $N$ is divisible by $c$. Our configuration uses $c=5$ and $N=10$: one token is committed every two Euler steps after the initial warm-up. A standard training state at a sampled phase $\tau$ is constructed directly from data,
\begin{equation}
  \vz_j(\tau)=\alpha_j(\tau)\vz^{\mathrm{GT}}_j+(1-\alpha_j(\tau))\vepsilon_j,
  \qquad \vepsilon_j\sim\mathcal{N}(\mathbf{0},\mathbf{I}),
  \label{eq:forward_path}
\end{equation}
and a bidirectional denoiser $v_\theta(\vz(\tau),\valpha(\tau),\vc)$ predicts velocities for the active tokens. An Euler update advances each position according to its actual change in denoising level,
\begin{equation}
  \vz^{k+1}_{j}=\vz^{k}_{j}+
  (\alpha^{k+1}_{j}-\alpha^{k}_{j})
  v_{\theta,j}(\vz^{k},\valpha^{k},\vc).
  \label{eq:euler}
\end{equation}
Once the leading token becomes clean, it is committed, the window shifts by one position, and fresh noise enters at the trailing edge. Repeating these operations extends generation beyond the training sequence length while retaining a bounded active window.

\section{Triangular Resampling}
\label{sec:method}

Standard FloodDiffusion training constructs each active window from ground-truth motion along the forward path in Equation~\ref{eq:forward_path}, whereas inference repeatedly updates, commits, and shifts model-generated states. The model therefore trains on ground-truth-derived windows but conditions on accumulated prediction errors during long rollouts.

This mismatch affects both completed history and partially denoised states within the active window. We propose \method{} to replay both regions (Figure~\ref{fig:overview}), with GT clamp controlling their exposure to model errors. Below-threshold states follow the noise-matched ground-truth path; states reaching the threshold retain model predictions. Replay construction, threshold control, and optimization are separate components: the supervised variant retains the original clean-motion target, while TR-DMD applies distribution matching to clean predictions read out from the replay.

\subsection{GT-Clamped Triangular Rollout}
\label{sec:replay}
\label{sec:clamp}

Each post-training sample with a nonempty history enters replay with probability $\gamma$, the \emph{resampling ratio}. For a replayed sample, we select an interval ending at the current output band and beginning a bounded number of tokens earlier, reset this interval to Gaussian noise at zero denoising progress, and keep any earlier prefix clean. The current model then follows the native triangular schedule through successive Euler updates with phase increment $\Delta\tau=1/N$ until the sampled training phase is reached, including the corresponding commits, window shifts, and fresh-noise injections. The resulting replay state is detached and passed to the subsequent optimization step.

For each replayed sample, we draw one release threshold using the shifted logit-normal parameterization~\citep{guo2025resampling}:
\begin{equation}
  r=\operatorname{sigmoid}(u+\log s),
  \qquad u\sim\mathcal{N}(0,1),
  \label{eq:threshold}
\end{equation}
where the shift $s>0$ controls clamp strength. The sigmoid maps an unconstrained Gaussian draw to a valid denoising threshold in $(0,1)$, while $\log s$ shifts its log-odds. This provides smooth control over the release point: smaller $s$ favors earlier release to model rollout, whereas larger $s$ retains ground-truth anchoring longer. The same $r$ is shared by all tokens and Euler steps in that replay. After each Euler update, token $j$ is replaced according to
\begin{equation}
  \widetilde{\vz}^{k+1}_{j}=
  \begin{cases}
    \alpha^{k+1}_{j}\vz^{\mathrm{GT}}_j+(1-\alpha^{k+1}_{j})\vepsilon^{k+1}_{j},
      & \alpha^{k+1}_{j}<r,\\
    \vz^{k+1}_{j}, & \alpha^{k+1}_{j}\geq r,
  \end{cases}
  \label{eq:clamp}
\end{equation}
with fresh $\vepsilon^{k+1}_{j}\sim\mathcal{N}(0,I)$ for the clamped branch. The clamp resamples the ground-truth latent at the location's current triangular noise level. As denoising progresses, a token crosses the common frontier and remains under model control.
Sharing the threshold preserves a contiguous denoising frontier across the window. The post-clamp state $\widetilde{\vz}^{k+1}$ becomes the input to the next Euler update.

The two hyperparameters have distinct roles. The ratio $\gamma$ determines how often optimization sees a replayed sample. The shift $s$ determines how much of an entered replay is anchored to ground truth. Small $s$ releases tokens early and approaches free rollout; large $s$ keeps more states clamped. TR-off is defined by $\gamma=0$. A fixed $s$ specifies a fixed threshold distribution.

\subsection{Optimization Objectives}
\label{sec:objective}

The replay construction specifies the training states, independently of the objective applied to model predictions. We use it with clean-motion supervision in \shortmethod{} and with distribution matching in TR-DMD.

\paragraph{GT-Supervised TR.}

For a training example sampled at phase $\tau$, let $\widetilde{\vz}$ denote the detached latent sequence left by the triangular replay of Section~\ref{sec:replay}, and let $\mathcal{B}(\tau)$ denote the output band supervised at this step. Replay ends at phase $\tau$, so the entries of $\widetilde{\vz}$ sit at the denoising levels $\valpha(\tau)$ prescribed by Equation~\ref{eq:triangular_schedule}. The optimization forward pass directly takes this replayed state as input, retaining its model-induced errors. For the linear path, we recover an effective noise from the replay state and the paired clean latent,
\begin{equation}
  \widehat{\vepsilon}_{j}=
  \frac{\widetilde{\vz}_{j}-\alpha_j(\tau)\vz^{\mathrm{GT}}_j}
       {\max(1-\alpha_j(\tau),\epsilon)},
  \qquad
  \vv^{\star}_{j}=\vz^{\mathrm{GT}}_j-\widehat{\vepsilon}_{j},
  \label{eq:recovery_target}
\end{equation}
where $\epsilon$ is a small numerical floor.
When $1-\alpha_j(\tau)\geq\epsilon$, Equation~\ref{eq:recovery_target} recovers the sampled noise for clamped tokens, giving the standard flow-matching target. Under the same condition, it defines the velocity that carries a model-generated state to the paired clean latent. The loss is
\begin{equation}
  \mathcal{L}_{\mathrm{TR}}(\theta)=
  \mathbb{E}\!\left[
  \frac{1}{|\mathcal{B}(\tau)|}
  \sum_{j\in\mathcal{B}(\tau)}
  \left\|v_{\theta,j}(\widetilde{\vz},\valpha(\tau),\vc)
  -\vv^{\star}_{j}\right\|_2^2\right].
  \label{eq:loss}
\end{equation}
Supervision is applied to $\mathcal{B}(\tau)$, while the remaining tokens provide context; the entire replayed window is treated as a fixed input. The objective therefore teaches the model how to continue denoising toward the paired clean motion when its own rollout displaces the active window. Samples that bypass replay use the ordinary forward-path state of Equation~\ref{eq:forward_path}, for which Equation~\ref{eq:loss} recovers the original FloodDiffusion objective.

\paragraph{Distribution Matching with TR-DMD.}
\label{sec:tr-dmd}

TR-DMD changes the optimization target while retaining triangular Euler replay and GT clamp. Propagated replay states remain detached. For each sample, we select one discrete denoising stage uniformly from the updates at which the threshold permits model control. When latent $j$ reaches that stage, its input state is saved and its velocity is recomputed with gradients, yielding the clean prediction
\begin{equation}
 \widehat{\vz}_{0,j}=\operatorname{sg}(\vz_{\alpha_j,j})
 +(1-\alpha_j)v_{\theta,j}\bigl(\operatorname{sg}(\vz_{\valpha}),\valpha,\vc\bigr),
 \label{eq:dmd_readout}
\end{equation}
where $\operatorname{sg}$ stops gradients. The recomputed clean predictions form the samples used for distribution matching. We optimize these predictions with a sampled-stage truncated-gradient approximation, treating the replayed states as fixed inputs.

Following the DMD update recipe of Rolling Forcing~\citep{liu2026rolling}, a frozen real-score model and a trainable fake-score model define a normalized, detached distribution-matching direction for the generator. The fake score learns a flow-matching objective on newly sampled, detached generator outputs. Both scores independently initialize from the same motion-teacher EMA; the generator initializes from a supervised \shortmethod{} checkpoint. GT anchors the replay states, while DMD supplies the generator's training objective. For distribution matching, we use up to the final 35 valid latents of each generated sequence (Appendix~\ref{app:tr-dmd-scoring}).

\section{Experiments}
\label{sec:experiments}

We organize the evaluation around four research questions. RQ1: Does controlled triangular rollout improve long-horizon motion quality and text alignment (Section~\ref{sec:main-results})? RQ2: How do \shortmethod{} and TR-DMD compare with existing methods within their respective objective groups (Section~\ref{sec:main-results})? RQ3: How do clamp strength, release structure, replay ratio, and threshold curriculum affect generation (Section~\ref{sec:ablations})? RQ4: How do blinded pairwise video preferences complement the quantitative metrics (Section~\ref{sec:human-evaluation})?

\subsection{Experimental Settings}
\label{sec:experimental-settings}

\paragraph{Datasets.}
We train on HumanML3D~\citep{guo2022humanml3d} and BABEL~\citep{punnakkal2021babel}. Both datasets use the standard 263-dimensional HumanML3D motion representation at 20\,fps. Evaluation uses prompts from the official HumanML3D test split.

\paragraph{Evaluation Protocol.}
Prior long-horizon motion evaluation assesses action segments and transitions~\citep{shafir2024prior,barquero2024flowmdm}, or uses sliding windows to evaluate extended sequences~\citep{lee2024t2lm}. We focus on how motion quality and text alignment evolve during sustained generation under a fixed text instruction. We generate 120-second sequences for 256 frozen HumanML3D test prompts and divide each sequence into twelve non-overlapping 10-second windows. At each window position, we compute FID, Matching Distance, and R-precision using the standard HumanML3D evaluator~\citep{guo2022humanml3d}. We report the resulting temporal curves, their normalized area under the curve (AUC), and linear degradation slopes. AUC summarizes performance across the full horizon, while slopes capture its temporal trend. We use FID AUC as the primary criterion for overall generation quality; the remaining metrics provide complementary evidence on text alignment and temporal degradation.

\paragraph{Training Setup.}
We initialize motion generators from the official FloodDiffusion checkpoint and use its corresponding pretrained VAE. TR-off and \shortmethod{} post-train for 30,000 optimization steps with the same data order, optimizer, learning-rate schedule, batch size, and single training seed. Each supervised post-training run uses one NVIDIA H200 GPU. Cumulative run times are approximately 21 hours for TR-off and 29 hours for the main \shortmethod{} configuration ($\gamma=0.25$, $s=0.6$). TR-DMD initializes its generator from the 30k supervised \shortmethod{} checkpoint at $(\gamma,s)=(1,0.6)$ and retains these replay settings during distribution matching. Its additional distribution-matching stage runs on two NVIDIA H200 GPUs for approximately 13 hours, performing 1,200 outer iterations with 1,200 fake-score updates and 240 generator updates. Additional implementation details are provided in Appendix~\ref{app:tr-implementation}.

\paragraph{Baselines.}
We select baselines to cover complementary strategies for addressing train--inference mismatch. TR-off and Gaussian history noise provide same-backbone controls for post-training and generic history corruption.
DART~\citep{zhao2025dart} and MotionStreamer~\citep{xiao2025motionstreamer} represent motion-domain training with model-induced histories. DART generates motion primitives, while MotionStreamer generates individual latents. Resampling Forcing~\citep{guo2025resampling} provides a teacher-free, supervised self-resampling alternative. Self Forcing~\citep{huang2025selfforcing}, Self Gradient Forcing~\citep{zhuang2026selfgradient}, Rolling Forcing~\citep{liu2026rolling}, and Causal Forcing~\citep{zhu2026causalforcing} initialization followed by Rolling Forcing provide video-derived DMD baselines. These video-derived methods use multi-latent chunks as autoregressive units. In contrast, our FloodDiffusion backbone jointly denoises an active window at staggered noise levels and commits one latent at a time. Adaptation protocols and training budgets are detailed in Appendix~\ref{app:external-protocols}.

\subsection{Main Results}
\label{sec:main-results}
\label{sec:external-comparison}

\begin{figure}[t]
  \centering
  \includegraphics[width=\linewidth]{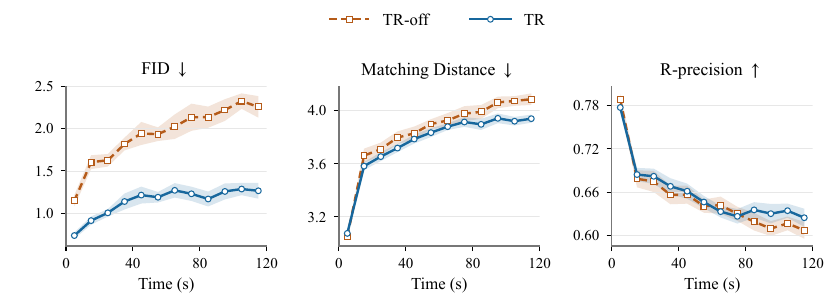}
  \caption{Window-wise quality and text alignment for TR-off and \shortmethod{} with $(\gamma,s)=(0.25,0.6)$. Each point summarizes a non-overlapping 10-second window over 256 prompts. Curves show means and shaded 95\% confidence intervals, under the convention in Table~\ref{tab:main}. Lower FID and Matching Distance and higher R-precision indicate better performance.}
  \label{fig:curves}
\end{figure}

Table~\ref{tab:main} reports the \shortmethod{} configuration selected by FID AUC on the same evaluation set. It improves over TR-off with the same backbone and number of post-training updates. FID AUC decreases from 1.951 to 1.153 (40.9\%), and the FID degradation slope falls from 0.526 to 0.235 per minute. Matching Distance AUC improves from 3.864 to 3.786, while R-precision AUC increases from 0.648 to 0.655. Although TR-DMD improves FID AUC over TR-off, it does not improve on its supervised $(\gamma,s)=(1,0.6)$ initialization (1.251 in Table~\ref{tab:ablation}); compatibility with DMD therefore does not imply an additional fidelity gain over supervised \shortmethod{}. Figure~\ref{fig:curves} shows how the supervised \shortmethod{}--TR-off differences evolve over time.

\begin{table}[htbp]
  \caption{Main comparisons grouped by without or with DMD. Following HumanML3D~\citep{guo2022humanml3d}, entries report means with 95\% confidence intervals after 20 rounds of evaluations. Slopes are per minute, with R-precision slopes in percentage points.}
  \label{tab:main}\label{tab:external}
  \centering
    \fontsize{7.5}{9.5}\selectfont
    \setlength{\tabcolsep}{0.5pt}
    \resizebox{\linewidth}{!}{%
    \begin{tabular}{@{}lcccccc@{}}
      \toprule
      \multirow{2}{*}[-0.4ex]{Method} & \multicolumn{2}{c}{FID} & \multicolumn{2}{c}{Matching Distance} & \multicolumn{2}{c}{R-precision} \\
      \cmidrule(lr){2-3}\cmidrule(lr){4-5}\cmidrule(lr){6-7}
       & AUC$\downarrow$ & slope$\downarrow$ & AUC$\downarrow$ & slope$\downarrow$ & AUC$\uparrow$ & slope$\uparrow$ \\
      \midrule
      \multicolumn{7}{l}{\textit{Without DMD}}\\[2pt]
      \shortmethod{} ($\gamma=0.25$, $s=0.6$) & \textbf{\boldmath\metricci{1.153}{0.030}} & \metricci{0.235}{0.035} & \metricci{3.786}{0.013} & \metricci{0.334}{0.015} & \metricci{0.655}{0.005} & \metricci{-5.804}{0.485} \\ 
      TR-off~\citep{cai2025flooddiffusion} & \metricci{1.951}{0.068} & \metricci{0.526}{0.070} & \metricci{3.864}{0.026} & \metricci{0.399}{0.028} & \metricci{0.648}{0.007} & \metricci{-6.837}{0.574} \\ 
      Gaussian noise ($\sigma=0.05$) & \metricci{\underline{1.410}}{0.035} & \metricci{0.306}{0.052} & \metricci{3.847}{0.020} & \metricci{0.316}{0.030} & \metricci{0.656}{0.006} & \metricci{-5.663}{0.597} \\ 
      DART~\citep{zhao2025dart} & \metricci{13.192}{0.134} & \metricci{-2.687}{0.266} & \metricci{5.300}{0.003} & \metricci{-0.080}{0.023} & \metricci{0.418}{0.004} & \metricci{1.276}{1.154} \\ 
      MotionStreamer~\citep{xiao2025motionstreamer} & \metricci{7.376}{0.436} & \metricci{3.363}{0.064} & \metricci{5.771}{0.097} & \metricci{0.812}{0.027} & \metricci{0.385}{0.011} & \metricci{-11.533}{0.990} \\ 
      Resampling Forcing~\citep{guo2025resampling} & \metricci{3.407}{0.140} & \metricci{1.573}{0.154} & \metricci{3.929}{0.008} & \metricci{0.475}{0.037} & \metricci{0.650}{0.009} & \metricci{-7.307}{1.457} \\ 
      \midrule
      \multicolumn{7}{l}{\textit{With DMD}}\\[2pt]
      TR-DMD ($\gamma=1$, $s=0.6$) & \textbf{\boldmath\metricci{1.324}{0.051}} & \metricci{0.442}{0.068} & \metricci{3.785}{0.018} & \metricci{0.270}{0.006} & \metricci{0.633}{0.012} & \metricci{-4.750}{0.282} \\ 
      Self Forcing~\citep{huang2025selfforcing} & \metricci{1.520}{0.086} & \metricci{-0.020}{0.118} & \metricci{3.566}{0.009} & \metricci{0.040}{0.053} & \metricci{0.690}{0.019} & \metricci{-0.284}{1.083} \\ 
      Self Gradient Forcing~\citep{zhuang2026selfgradient} & \metricci{4.005}{0.435} & \metricci{0.322}{0.279} & \metricci{4.507}{0.123} & \metricci{0.282}{0.094} & \metricci{0.541}{0.028} & \metricci{-3.182}{2.008} \\ 
      Rolling Forcing~\citep{liu2026rolling} & \metricci{\underline{1.351}}{0.061} & \metricci{0.026}{0.097} & \metricci{3.574}{0.018} & \metricci{-0.010}{0.037} & \metricci{0.694}{0.017} & \metricci{0.270}{0.454} \\ 
      \begin{tabular}[c]{@{}l@{}}Causal Forcing Init.~\citep{zhu2026causalforcing}\\+ Rolling Forcing\end{tabular} & \metricci{1.634}{0.035} & \metricci{0.116}{0.102} & \metricci{3.199}{0.033} & \metricci{0.000}{0.055} & \metricci{0.749}{0.007} & \metricci{-0.268}{0.600} \\ 
      \bottomrule
    \end{tabular}}
\end{table}

Without DMD, \shortmethod{} achieves the best FID AUC (1.153). Even TR-off already outperforms DART, MotionStreamer, and Resampling Forcing in both FID AUC (1.951 versus 13.192, 7.376, and 3.407) and Matching Distance AUC (3.864 versus 5.300, 5.771, and 3.929). This strong baseline supports the effectiveness of FloodDiffusion's triangular, latent-wise denoising strategy for long-horizon generation under our protocol. Gaussian history noise further reduces FID AUC to 1.410 and Matching Distance AUC to 3.847, making it a competitive same-backbone baseline. Nevertheless, \shortmethod{} improves these scores to 1.153 and 3.786, respectively, including an 18.2\% reduction in FID AUC relative to Gaussian noise. Structured triangular replay therefore yields greater fidelity gains than generic history corruption in this controlled comparison.

With DMD, TR-DMD achieves the best FID AUC (1.324), followed by Rolling Forcing (1.351), Self Forcing (1.520), Causal Forcing initialization followed by Rolling Forcing (1.634), and Self Gradient Forcing (4.005). Rolling Forcing~\citep{liu2026rolling} likewise uses staggered noise levels, but operates on multi-latent chunks rather than committing individual latents. Its second-place result, together with the strong \shortmethod{} results, is consistent with the benefit of staggered denoising schedules for long-horizon motion fidelity.

\subsection{Ablation Study}
\label{sec:ablations}

\begin{table}[htbp]
  \caption{Triangular Resampling ablations. Entries follow the mean and 95\% confidence-interval convention of Table~\ref{tab:main}. Bold rows mark the lowest FID AUC within each group.}
  \label{tab:ablation}\label{tab:mechanisms}
  \centering
    \fontsize{8}{10}\selectfont
    \setlength{\tabcolsep}{1pt}
    \renewcommand{\arraystretch}{1.12}
    \resizebox{\linewidth}{!}{%
    \begin{tabular}{@{}l@{\hspace{4pt}}l@{\hspace{4pt}}cccccc@{}}
      \toprule
      \multirow{2}{*}[-0.4ex]{Group} & \multirow{2}{*}[-0.4ex]{Setting} & \multicolumn{2}{c}{FID} & \multicolumn{2}{c}{Matching Distance} & \multicolumn{2}{c}{R-precision} \\
      \cmidrule(lr){3-4}\cmidrule(lr){5-6}\cmidrule(lr){7-8}
       & & AUC$\downarrow$ & slope$\downarrow$ & AUC$\downarrow$ & slope$\downarrow$ & AUC$\uparrow$ & slope$\uparrow$ \\
      \midrule
      \multirow{11}{*}{\shortstack[l]{Shift $s$\\($\gamma=1$)}} & 0 (full self-rollout) & \metricci{16.024}{0.064} & \metricci{1.151}{0.065} & \metricci{6.189}{0.006} & \metricci{0.240}{0.008} & \metricci{0.357}{0.004} & \metricci{-3.935}{0.408} \\ 
       & 0.3 & \metricci{2.044}{0.047} & \metricci{0.459}{0.062} & \metricci{3.853}{0.020} & \metricci{0.285}{0.014} & \metricci{0.631}{0.005} & \metricci{-4.335}{0.354} \\ 
       & \textbf{0.6} & \textbf{\boldmath\metricci{1.251}{0.024}} & \textbf{\boldmath\metricci{0.376}{0.039}} & \textbf{\boldmath\metricci{3.780}{0.014}} & \textbf{\boldmath\metricci{0.259}{0.021}} & \textbf{\boldmath\metricci{0.646}{0.005}} & \textbf{\boldmath\metricci{-4.391}{0.492}} \\ 
       & 1 & \metricci{1.924}{0.054} & \metricci{0.650}{0.063} & \metricci{4.020}{0.019} & \metricci{0.417}{0.018} & \metricci{0.639}{0.006} & \metricci{-6.103}{0.403} \\ 
       & 3 & \metricci{1.349}{0.036} & \metricci{0.300}{0.060} & \metricci{3.663}{0.011} & \metricci{0.287}{0.029} & \metricci{0.679}{0.004} & \metricci{-5.041}{0.335} \\ 
       & 10 & \metricci{1.983}{0.063} & \metricci{0.425}{0.084} & \metricci{3.659}{0.018} & \metricci{0.240}{0.031} & \metricci{0.681}{0.006} & \metricci{-4.036}{0.391} \\ 
       & 100 & \metricci{2.765}{0.072} & \metricci{0.967}{0.089} & \metricci{3.819}{0.018} & \metricci{0.354}{0.025} & \metricci{0.654}{0.005} & \metricci{-5.837}{0.627} \\ 
       & 1000 & \metricci{2.780}{0.095} & \metricci{1.004}{0.158} & \metricci{3.832}{0.023} & \metricci{0.375}{0.036} & \metricci{0.654}{0.005} & \metricci{-6.067}{0.638} \\ 
       & 10000 & \metricci{2.780}{0.095} & \metricci{1.004}{0.158} & \metricci{3.832}{0.023} & \metricci{0.375}{0.036} & \metricci{0.654}{0.005} & \metricci{-6.067}{0.638} \\ 
       & Random-position clamp & \metricci{1.857}{0.052} & \metricci{0.510}{0.052} & \metricci{3.841}{0.018} & \metricci{0.374}{0.023} & \metricci{0.643}{0.004} & \metricci{-6.255}{0.602} \\ 
       & Curriculum ($3\rightarrow0.6$) & \metricci{1.455}{0.015} & \metricci{0.348}{0.025} & \metricci{3.877}{0.034} & \metricci{0.287}{0.020} & \metricci{0.638}{0.010} & \metricci{-5.100}{0.641} \\ 
      \midrule
      \multirow{3}{*}{\shortstack[l]{Ratio $\gamma$\\($s=0.6$)}} & \textbf{0.25} & \textbf{\boldmath\metricci{1.153}{0.030}} & \textbf{\boldmath\metricci{0.235}{0.035}} & \textbf{\boldmath\metricci{3.786}{0.013}} & \textbf{\boldmath\metricci{0.334}{0.015}} & \textbf{\boldmath\metricci{0.655}{0.005}} & \textbf{\boldmath\metricci{-5.804}{0.485}} \\ 
       & 0.5 & \metricci{1.495}{0.023} & \metricci{0.438}{0.049} & \metricci{3.929}{0.014} & \metricci{0.349}{0.024} & \metricci{0.622}{0.004} & \metricci{-6.038}{0.486} \\ 
       & 1 & \metricci{1.251}{0.024} & \metricci{0.376}{0.039} & \metricci{3.780}{0.014} & \metricci{0.259}{0.021} & \metricci{0.646}{0.005} & \metricci{-4.391}{0.492} \\ 
      \bottomrule
    \end{tabular}}
\end{table}

All ablations start from the official pretrained FloodDiffusion checkpoint and use 30k post-training updates. Table~\ref{tab:ablation} varies the threshold shift $s$ at $\gamma=1$ and the replay probability $\gamma$ at $s=0.6$, with contiguous-frontier clamping by default. Random-position clamping uses $(\gamma,s)=(1,0.6)$ and uniformly redistributes the frontier mask's exact clamp count over the replay region at each Euler step. Curriculum fixes $\gamma=1$, cosine-interpolates $\log s$ from $\log 3$ to $\log 0.6$ over the first 15k updates, and holds $s=0.6$ for the remaining 15k. Full self-rollout, labeled 0, disables GT clamping rather than setting $s=0$.

For threshold control, full self-rollout severely degrades motion quality despite the pretrained initialization: its FID AUC reaches 16.024, compared with 1.251 for GT-clamped replay at $s=0.6$. R-precision AUC also falls from 0.646 to 0.357, demonstrating the importance of GT anchoring in this setting. The shift sweep identifies an empirical optimum at $s=0.6$ among the tested values: FID AUC is 2.044 at $s=0.3$, 1.349 at $s=3$, and at least 2.765 for $s\geq100$. For $s=1000$ and $s=10000$, thresholds concentrate near 1, keeping nearly all intermediate denoising states clamped to noise-matched GT. Both settings thus approach GT-based training and produce identical evaluation results in our runs. Random-position clamping retains much of the benefit of GT anchoring, reaching FID AUC 1.857, while curriculum performs better at 1.455. Neither matches the fixed $s=0.6$ threshold distribution (1.251), indicating that a well-chosen fixed shift with contiguous release is sufficient for the best observed fidelity in this sweep.

For replay probability, the tested nonzero ratios show a non-monotonic pattern: both infrequent replay ($\gamma=0.25$) and replay on every eligible sample ($\gamma=1$) outperform the intermediate mixture ($\gamma=0.5$). Their FID AUCs are 1.153, 1.251, and 1.495, respectively, with corresponding degradation slopes of 0.235, 0.376, and 0.438. Thus, within this grid, either lower or higher replay frequency is preferable to an equal mixture of replayed and ordinary training samples; $\gamma=0.25$ gives the best FID AUC and slope.

\FloatBarrier
\subsection{Human Evaluation}
\label{sec:human-evaluation}

To assess the overall visual quality of generated motion, we conduct a pairwise preference test. We compare \shortmethod{}, the matched TR-off (FloodDiffusion) baseline, and Rolling Forcing using 30 prompts, with one generated motion per method and prompt. For each prompt, all three method pairs are presented side by side in three rounds, with method identities hidden and left--right order varied across rounds. Each comparison yields separate choices for motion quality and text alignment. We aggregate all 270 pairwise choices per criterion into Bradley--Terry (BT) scores. Appendix~\ref{app:human-evaluation} provides the protocol and complete prompt list.

Table~\ref{tab:human_evaluation} shows that \shortmethod{} achieves the highest BT scores for both motion quality and text alignment. For motion quality, \shortmethod{} scores 0.302, compared with $-0.272$ for TR-off and $-0.030$ for Rolling Forcing. For text alignment, the corresponding scores are 0.317, $-0.211$, and $-0.106$. Both criteria yield the same ranking, with a larger score gap between \shortmethod{} and TR-off than between \shortmethod{} and Rolling Forcing. This consistent ranking indicates that \shortmethod{} is preferred for both motion quality and text alignment among the compared methods.

\begin{table}[!ht]
  \centering\small
  \setlength{\tabcolsep}{3pt}
  \caption{Pairwise video preferences on 30 prompts. Three rounds provide three judgments per pair and prompt (270 comparisons), with separate choices for motion quality and text alignment. Entries are zero-mean log-strength Bradley--Terry scores fitted jointly across all three rounds; higher is better. Bold marks the highest score in each column.}
  \label{tab:human_evaluation}
  \begin{tabular*}{\linewidth}{@{\extracolsep{\fill}}lcc@{}}
    \toprule
    Method & Motion Quality $\uparrow$ & Text Alignment $\uparrow$ \\
    \midrule
    \shortmethod{} & $\mathbf{0.302}$ & $\mathbf{0.317}$ \\
    TR-off~\citep{cai2025flooddiffusion} & $-0.272$ & $-0.211$ \\
    Rolling Forcing~\citep{liu2026rolling} & $-0.030$ & $-0.106$ \\
    \bottomrule
  \end{tabular*}
\end{table}

\FloatBarrier
\section{Conclusions}
\label{sec:conclusions}

We propose Triangular Resampling (TR) to improve long-horizon motion generation under triangular denoising. TR replays the active-window trajectory, including partially denoised states, and uses ground-truth clamping to control their release into model rollouts. Supervised TR reduces FID AUC by 40.9\% relative to matched TR-off. TR and TR-DMD attain the lowest mean FID AUC within their respective comparison groups under this protocol. Ablations support controlled GT anchoring over full self-rollout or random-position clamping, while TR-DMD demonstrates compatibility with distribution matching. These results highlight the importance of train-inference match in the active denoising area beyond the completed history. We leave extensions to other backbones, denoising schedules, and post-training data settings to future work.

\paragraph{Limitations.}
Multi-step replay slows post-training and increases its computational cost. In our initial 5k-update runs on shared GPUs, \shortmethod{} at $\gamma=0.25$ takes approximately 1.95 seconds per update, compared with 0.58 seconds for TR-off (about $3.3\times$). Shorter stochastic rollouts combined with sampled-step supervision and gradient truncation, inspired by Self Forcing~\citep{huang2025selfforcing}, may help reduce this overhead.

\section*{Ethics Statement}
We use existing motion datasets and pretrained models in accordance with their respective licenses and access terms. Our human evaluation respects participants' rights, dignity, and privacy. Participation is voluntary and based on informed consent, and participants can withdraw at any time. Participants receive compensation at an effective hourly rate above the local average hourly wage. All responses are de-identified and used solely for research purposes.

\section*{Reproducibility Statement}
Equations~\ref{eq:triangular_schedule}--\ref{eq:dmd_readout} specify the schedule, replay, clamp distribution, supervised target, and differentiable DMD readout. Section~\ref{sec:experiments} records the evaluation horizon, windowing, supervised and DMD update budgets, initialization, and ablation grid. Appendix~\ref{app:external-protocols} records the external-method training and inference protocols. We will publicly release the implementation, training configurations, frozen prompt manifests, checkpoint identifiers, and evaluation scripts to support reproducibility.

\section*{AI Use Statement}
Generative AI tools were used to assist with literature discovery and summarization, experiment-planning feedback, software implementation and debugging, and drafting and editing portions of this manuscript and bibliography. The authors reviewed the cited primary sources, technical descriptions, code behavior, and all AI-assisted text, and take responsibility for the final content, claims, and artifacts of the submission.

\bibliography{iclr2027_conference}

@inproceedings{athanasiou2022teach,
  title         = {{TEACH}: Temporal Action Composition for 3D Humans},
  author        = {Athanasiou, Nikos and Petrovich, Mathis and Black, Michael J. and Varol, G{\"u}l},
  booktitle     = {International Conference on 3D Vision},
  pages         = {414--423},
  year          = {2022}
}

@inproceedings{barquero2024flowmdm,
  title     = {Seamless Human Motion Composition with Blended Positional Encodings},
  author    = {Barquero, German and Escalera, Sergio and Palmero, Cristina},
  booktitle = {Proceedings of the IEEE/CVF Conference on Computer Vision and Pattern Recognition},
  pages     = {457--469},
  year      = {2024}
}

@inproceedings{cai2025flooddiffusion,
  title         = {{FloodDiffusion}: Tailored Diffusion Forcing for Streaming Motion Generation},
  author        = {Cai, Yiyi and Wu, Yuhan and Li, Kunhang and Zhou, You and Zheng, Bo and Liu, Haiyang},
  booktitle     = {Proceedings of the IEEE/CVF Conference on Computer Vision and Pattern Recognition},
  pages         = {2295--2304},
  year          = {2026}
}

@inproceedings{guo2022humanml3d,
  title     = {Generating Diverse and Natural 3D Human Motions From Text},
  author    = {Guo, Chuan and Zou, Shihao and Zuo, Xinxin and Wang, Sen and Ji, Wei and Li, Xingyu and Cheng, Li},
  booktitle = {Proceedings of the IEEE/CVF Conference on Computer Vision and Pattern Recognition},
  pages     = {5152--5161},
  year      = {2022}
}

@article{guo2025resampling,
  title         = {End-to-End Training for Autoregressive Video Diffusion via Self-Resampling},
  author        = {Guo, Yuwei and Yang, Ceyuan and He, Hao and Zhao, Yang and Wei, Meng and Yang, Zhenheng and Huang, Weilin and Lin, Dahua},
  journal       = {arXiv preprint arXiv:2512.15702},
  year          = {2025},
  eprint        = {2512.15702},
  archivePrefix = {arXiv}
}

@inproceedings{huang2025selfforcing,
  title     = {Self Forcing: Bridging the Train-Test Gap in Autoregressive Video Diffusion},
  author    = {Huang, Xun and Li, Zhengqi and He, Guande and Zhou, Mingyuan and Shechtman, Eli},
  booktitle = {Advances in Neural Information Processing Systems},
  volume    = {38},
  pages     = {167283--167308},
  doi       = {10.52202/085713-5576},
  year      = {2025}
}

@inproceedings{lee2024t2lm,
  title     = {{T2LM}: Long-Term 3D Human Motion Generation from Multiple Sentences},
  author    = {Lee, Taeryung and Baradel, Fabien and Lucas, Thomas and Lee, Kyoung Mu and Rogez, Gr{\`e}gory},
  booktitle = {Proceedings of the IEEE/CVF Conference on Computer Vision and Pattern Recognition Workshops},
  pages     = {1867--1876},
  year      = {2024}
}

@article{ling2026prismv1,
  title         = {{PRISM}: Streaming Human Motion Generation with Per-Joint Latent Decomposition},
  author        = {Ling, Zeyu and Shuai, Qing and Zhang, Teng and Li, Shiyang and Han, Bo and Zou, Changqing},
  journal       = {arXiv preprint arXiv:2603.08590v1},
  year          = {2026},
  eprint        = {2603.08590v1},
  archivePrefix = {arXiv},
  note          = {Version 1}
}

@inproceedings{lipman2023flowmatching,
  title     = {Flow Matching for Generative Modeling},
  author    = {Lipman, Yaron and Chen, Ricky T. Q. and Ben-Hamu, Heli and Nickel, Maximilian and Le, Matthew},
  booktitle = {International Conference on Learning Representations},
  year      = {2023}
}

@inproceedings{liu2026rolling,
  title     = {Rolling Forcing: Autoregressive Long Video Diffusion in Real Time},
  author    = {Liu, Kunhao and Hu, Wenbo and Xu, Jiale and Shan, Ying and Lu, Shijian},
  booktitle = {International Conference on Learning Representations},
  pages     = {91177--91196},
  year      = {2026}
}

@inproceedings{punnakkal2021babel,
  title     = {{BABEL}: Bodies, Action and Behavior With English Labels},
  author    = {Punnakkal, Abhinanda R. and Chandrasekaran, Arjun and Athanasiou, Nikos and Quiros-Ramirez, Alejandra and Black, Michael J.},
  booktitle = {Proceedings of the IEEE/CVF Conference on Computer Vision and Pattern Recognition},
  pages     = {722--731},
  year      = {2021}
}

@inproceedings{shafir2024prior,
  title     = {Human Motion Diffusion as a Generative Prior},
  author    = {Shafir, Yonatan and Tevet, Guy and Kapon, Roy and Bermano, Amit H.},
  booktitle = {International Conference on Learning Representations},
  pages     = {8717--8733},
  year      = {2024}
}

@inproceedings{xiao2025motionstreamer,
  title     = {{MotionStreamer}: Streaming Motion Generation via Diffusion-based Autoregressive Model in Causal Latent Space},
  author    = {Xiao, Lixing and Lu, Shunlin and Pi, Huaijin and Fan, Ke and Pan, Liang and Zhou, Yueer and Feng, Ziyong and Zhou, Xiaowei and Peng, Sida and Wang, Jingbo},
  booktitle = {Proceedings of the IEEE/CVF International Conference on Computer Vision},
  pages     = {10086--10096},
  year      = {2025}
}

@inproceedings{yin2024onestep,
  title     = {One-step Diffusion with Distribution Matching Distillation},
  author    = {Yin, Tianwei and Gharbi, Micha{\"e}l and Zhang, Richard and Shechtman, Eli and Durand, Fr{\'e}do and Freeman, William T. and Park, Taesung},
  booktitle = {Proceedings of the IEEE/CVF Conference on Computer Vision and Pattern Recognition},
  pages     = {6613--6623},
  year      = {2024}
}

@inproceedings{zhao2025dart,
  title     = {{DartControl}: A Diffusion-Based Autoregressive Motion Model for Real-Time Text-Driven Motion Control},
  author    = {Zhao, Kaifeng and Li, Gen and Tang, Siyu},
  booktitle = {International Conference on Learning Representations},
  pages     = {23569--23592},
  year      = {2025}
}

@inproceedings{zhu2026causalforcing,
  title         = {Causal Forcing: Autoregressive Diffusion Distillation Done Right for High-Quality Real-Time Interactive Video Generation},
  author        = {Zhu, Hongzhou and Zhao, Min and He, Guande and Su, Hang and Li, Chongxuan and Zhu, Jun},
  booktitle     = {Proceedings of the International Conference on Machine Learning},
  year          = {2026}
}

@article{zhuang2026selfgradient,
  title         = {Self Gradient Forcing: Native Long Video Extrapolation},
  author        = {Zhuang, Junhao and Zhang, Shiyi and Bian, Yuxuan and Li, Yaowei and Luo, Yawen and Liu, Yijun and Jin, Weiyang and Zhang, Songchun and He, Xianglong and Zhang, Xuying and Li, Haoran and Huang, Haoyang and Xue, Zeyue and Duan, Nan},
  journal       = {arXiv preprint arXiv:2607.20368},
  year          = {2026},
  eprint        = {2607.20368},
  archivePrefix = {arXiv}
}
\bibliographystyle{iclr2027_conference}

\clearpage
\appendix
\section{TR and TR-DMD Implementation Details}
\label{app:tr-implementation}
\label{app:tr-dmd-scoring}

\paragraph{Supervised TR.}
Checkpoints are saved every 5,000 optimization steps, and formal comparisons use the final 30k checkpoint. The main supervised \shortmethod{} uses $(\gamma,s)=(0.25,0.6)$ and contiguous-frontier clamping, the lowest-FID-AUC configuration in the reported ratio sweep (Table~\ref{tab:ablation}). This hyperparameter choice uses the reported evaluations, not a held-out selection set.

\paragraph{TR-DMD Optimization.}
Each generator and fake-score update uses effective batch size 64. Generator and fake-score learning rates are $1.5\times10^{-6}$ and $4\times10^{-7}$, with AdamW betas $(0,0.999)$, no weight decay, score-time shift 5, gradient clipping at 10, and generator EMA decay 0.99 starting at iteration 200. The two scores independently load the same 10k-update motion-teacher EMA used by Rolling Forcing. We evaluate the final 1,200-iteration checkpoint. The total training budget includes the preceding 30k supervised updates.

\paragraph{DMD Scoring Window.}
The real-score and fake-score networks evaluate at most the final 35 latents of each generated sequence. Cropped sequences undergo causal-VAE boundary re-encoding, and shorter sequences are processed with their actual valid lengths.

\clearpage
\section{External-Method Training and Inference Protocols}
\label{app:external-protocols}
\setcounter{table}{0}
\renewcommand{\thetable}{B\arabic{table}}
\renewcommand{\theHtable}{appendix.B.\arabic{table}}

All comparison models use HumanML3D+BABEL training data and share frozen evaluation prompts, sequence duration, output frame rate, reference bank, and evaluator weights. Metrics are computed independently within each repeat rather than pooling embeddings. We adapt the external systems to 263-dimensional motion using released code and method-specific rollout objectives. The non-DMD and DMD groups in Table~\ref{tab:main} distinguish training objectives; teacher, initialization, rollout windows, effective batch sizes, and update budgets remain method-specific.

TR-off continues post-training without triangular replay under the same backbone and 30k update budget as supervised \shortmethod{}. Gaussian history noise adds $0.05\vepsilon$ independently to conditioning history while leaving the supervised band unchanged, with matched initialization, training data, and updates. It serves as a baseline comparison rather than a \shortmethod{} ablation.

Table~\ref{tab:external-protocols} records the motion-adaptation budgets and evaluated checkpoints. The video-derived students use a motion DiT and motion teacher, not the original video-scale models. The shared bidirectional teacher receives 10k updates; causal AR and ODE initialization stages each use 10k updates where applicable, with 4,096 ODE records. Self Forcing and Rolling Forcing start from the non-causal teacher's ODE distillation; Self Gradient Forcing starts from the causal AR model. For Causal Forcing, we follow the authors' released long-video extension: causal ODE initialization followed by Rolling Forcing training and inference. This is our chosen long-horizon baseline protocol, not a requirement of Causal Forcing itself. Its rolling stage matches our Rolling Forcing adaptation's five-step, five-latent configuration to compare initialization choices. These are fixed adaptation budgets, not claims of convergence at the original video scale.

An iteration denotes the outer DMD iteration, not one generator update. Self Forcing, Rolling Forcing, and the Causal Forcing adaptation use effective batch size 64; Self Gradient Forcing uses 8. Their blocks contain five motion latents (one second at 20\,fps), rather than the video implementations' three-latent blocks. NFE (number of function evaluations) counts denoising-network evaluations during inference. We report NFE per committed autoregressive unit: a motion primitive, a single latent token, or a multi-latent chunk, depending on the method. For example, ``4 per 5-latent unit'' means four denoising evaluations per generated chunk of five latents. MotionStreamer and Resampling Forcing use their predeclared final endpoints of 100k and 31.5k, respectively, without test-based checkpoint selection. The RF endpoint includes its full training curriculum, including the 15-second and history-routing stages.

\begin{table}[!ht]
  \caption{Training and inference protocols for the external 263-dimensional motion adaptations. ``AR'', ``ODE'', and ``DMD'' denote the causal autoregressive-teacher, ODE-initialization, and distribution-matching stages, respectively. NFE counts denoising-network evaluations per committed unit, with each method's unit stated explicitly.}
  \label{tab:external-protocols}
  \centering
  \scriptsize
  \resizebox{\linewidth}{!}{%
  \begin{tabular}{lll}
    \toprule
    Method & Training Endpoint & Inference NFE per Committed Unit \\
    \midrule
    DART~\citep{zhao2025dart} & VAE 200k; denoiser 300k & 10 per 8-frame primitive \\
    MotionStreamer~\citep{xiao2025motionstreamer} & TAE budget 2M; generator 100k & 50 per latent token \\
    Resampling Forcing~\citep{guo2025resampling} & Full curriculum 31.5k & 32 per 3-latent unit \\
    Self Forcing~\citep{huang2025selfforcing} & ODE 10k; DMD 1k iterations & 4 per 5-latent unit \\
    Self Gradient Forcing~\citep{zhuang2026selfgradient} & AR 10k; DMD 1.2k iterations & 4 per 5-latent unit \\
    Rolling Forcing~\citep{liu2026rolling} & ODE 10k; DMD 1.2k iterations & 5 per 5-latent unit \\
    \shortstack[l]{Causal Forcing Initialization~\citep{zhu2026causalforcing}\\+ Rolling Forcing} & AR 10k; ODE 10k; rolling DMD 1.2k & 5 per 5-latent unit \\
    \bottomrule
  \end{tabular}}
\end{table}

\clearpage
\section{Pairwise Preference Evaluation Protocol}
\label{app:human-evaluation}
\setcounter{table}{0}
\renewcommand{\thetable}{C\arabic{table}}
\renewcommand{\theHtable}{appendix.C.\arabic{table}}
\setcounter{figure}{0}
\renewcommand{\thefigure}{C\arabic{figure}}
\renewcommand{\theHfigure}{appendix.C.\arabic{figure}}

\paragraph{Methods and Cases.}
We compare \shortmethod{}, TR-off, and Rolling Forcing, adapting the pairwise designs of FloodDiffusion~\citep{cai2025flooddiffusion} and DART~\citep{zhao2025dart}. TR-off is the matched FloodDiffusion baseline post-trained for 30k updates without triangular replay, rather than the unadapted official checkpoint. \shortmethod{} uses the 30k $(\gamma,s)=(0.25,0.6)$ checkpoint, and Rolling Forcing uses the checkpoint evaluated in Table~\ref{tab:main}. We select 30 fixed-text prompts from the frozen 256-prompt HumanML3D evaluation manifest by reviewing the text for suitability for sustained motion. We retain ongoing or cyclic actions and exclude one-shot transitions, explicit stopping events, fixed-count or fixed-duration actions, and finite multi-event scripts. The set covers locomotion (10), running (5), dance (4), repetitive exercise (7), and gestures or ground motion (4). Selection uses the text only, not method outputs or metric scores. Table~\ref{tab:human-prompts} lists the prompts verbatim. Each method supplies one 120-second motion per prompt, reusing available motions from the quantitative evaluation or earlier review renders of the same checkpoints.

\begin{figure}[htbp]
  \centering
  {\small\makebox[0.49\linewidth]{A (left)}\hfill\makebox[0.49\linewidth]{B (right)}\par}
  \includegraphics[width=\linewidth]{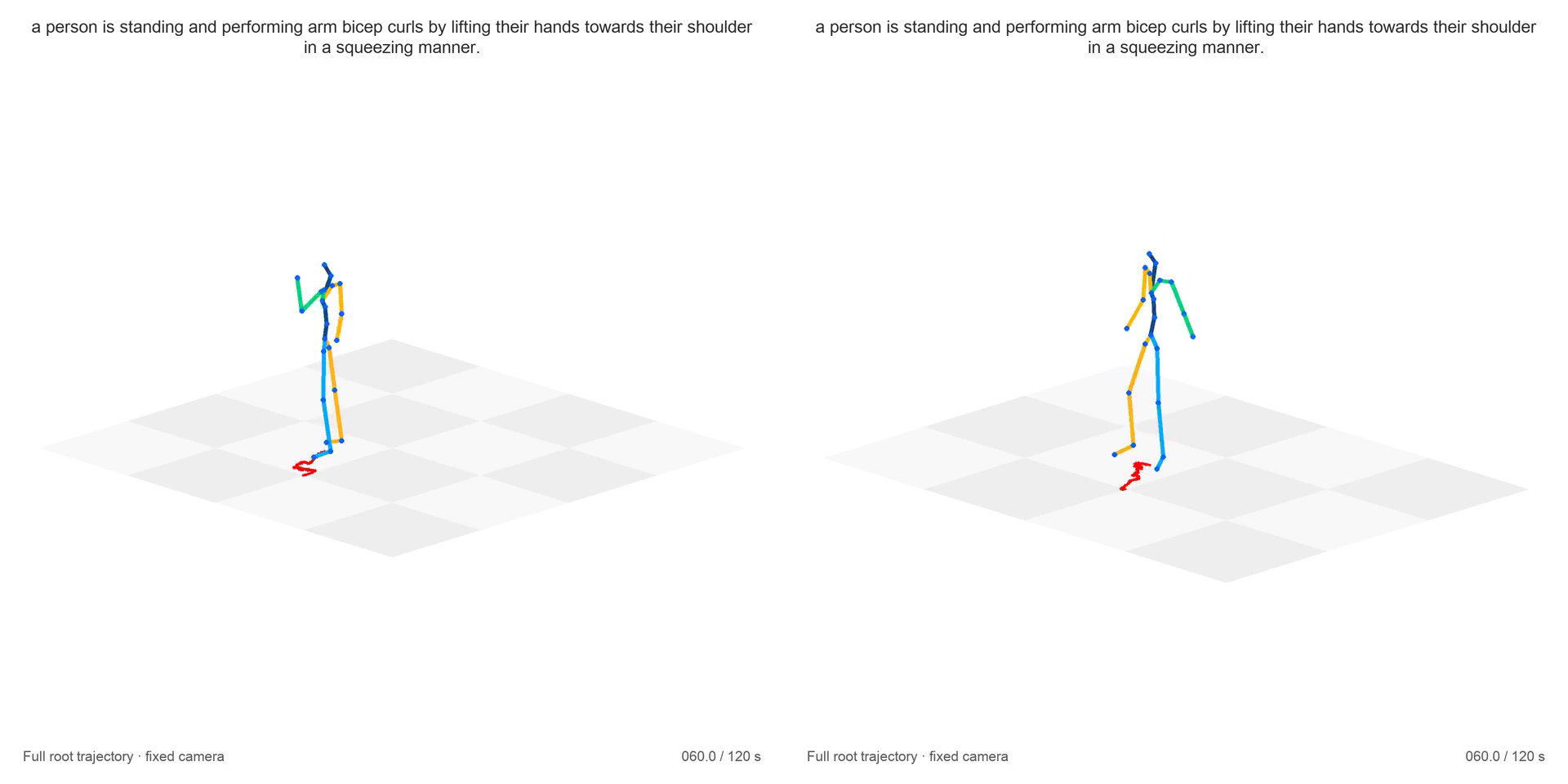}
  \caption{Example comparison video shown to annotators, captured at 60 seconds. A and B identify the left and right videos; method identities are hidden. The frame preserves the original camera views, skeleton colors, and root trajectories.}
  \label{fig:annotator-input}
\end{figure}

\paragraph{Rendering and Presentation.}
All 90 videos use the same 22-joint skeleton renderer, 960-by-960 resolution, 20\,fps, and FloodDiffusion's default 263D skeleton colors. Joint positions are recovered directly from the 263-dimensional motion features, without inverse kinematics, smoothing, or temporal resampling. A fixed camera frames each complete motion above a checkerboard floor; a red trail retains the entire ground-plane root trajectory up to the current frame. Videos contain the prompt and playback time but no method labels. Each pair is concatenated horizontally into one synchronized 120-second video, with the left and right halves identified as A and B. This presentation exposes global trajectory behavior. Figure~\ref{fig:annotator-input} illustrates the presentation with a frame at 60 seconds for prompt \texttt{000467}, selected for the visibility of both figures; annotators view the complete synchronized videos.

\paragraph{Pairing and Comparisons.}
The 90 videos yield 90 within-prompt pairs: \shortmethod{} versus TR-off, \shortmethod{} versus Rolling Forcing, and TR-off versus Rolling Forcing for each prompt. The reported results use all three completed rounds, yielding three judgments per pair and prompt (270 comparisons). Left--right order is pseudorandomized per pair and prompt in the first round, reversed in the second, and restored in the third; method identities remain hidden. All rounds assess the same videos.

\paragraph{Questions and Collection Rules.}
For each pair, annotators make separate forced A/B choices for overall motion quality and text alignment, yielding 540 binary responses across the 270 comparisons. Quality concerns natural, coherent movement without conspicuous jitter, foot sliding, freezing, or implausible poses; alignment concerns sustained adherence to the displayed instruction. There is no numerical rating or tie option. Annotators assess the complete video pair while method identities remain hidden.

\paragraph{Analysis.}
We fit a Bradley--Terry model separately for motion quality and text alignment, with $P(i\succ j)=\exp(b_i)/(\exp(b_i)+\exp(b_j))$. Unregularized maximum likelihood pools all 270 pairwise judgments per criterion across the three rounds (90 per method pair); we report the fitted log-strengths $b_i$, centered so that $\sum_i b_i=0$. We fit once to the pooled votes rather than averaging round-wise scores. Higher scores indicate stronger preference in the joint fit, not pairwise win percentages.

\FloatBarrier

\begingroup
\small
\setlength{\tabcolsep}{4pt}
\begin{longtable}{@{}rlp{0.73\linewidth}@{}}
\caption{The 30 prompts selected from the quantitative-evaluation manifest, reproduced verbatim.}\label{tab:human-prompts}\\
\toprule
No. & Prompt ID & Text \\
\midrule
\endfirsthead
\toprule
No. & Prompt ID & Text (Continued) \\
\midrule
\endhead
\bottomrule
\endfoot
1 & \texttt{000467} & a person is standing and performing arm bicep curls by lifting their hands towards their shoulder in a squeezing manner. \\
2 & \texttt{000556} & a person appears to be doing a dance. \\
3 & \texttt{000710} & a person walks forward while twisting their torso side to side. \\
4 & \texttt{001215} & a man sways side by side with arms out \\
5 & \texttt{001313} & a man walks forward while swaying his feet in a zig-zag path. \\
6 & \texttt{002848} & a man is pacing back and forth in a straight line. \\
7 & \texttt{003005} & he does a salsa dance \\
8 & \texttt{003020} & a man slowly sways from side to side, sightly bending his knees. \\
9 & \texttt{004822} & a person is walking in place at a slow pace. \\
10 & \texttt{005609} & a person jogs in place slowly in a counter clockwise circle. \\
11 & \texttt{006251} & the person is walking normally. \\
12 & \texttt{006523} & a person runs forward in a non-linear way. \\
13 & \texttt{007354} & he starts to crawl a lot \\
14 & \texttt{007418} & a person raises the arm and waves multiple times. \\
15 & \texttt{008100} & a person walks in a counter clockwise circle. \\
16 & \texttt{008296} & the man dances his feet in circles in front of himself. \\
17 & \texttt{009478} & a person walks backwards in a straight line \\
18 & \texttt{009561} & a person runs fast diagonal. \\
19 & \texttt{010114} & person is making stepping motion in place \\
20 & \texttt{011397} & a man walks clockwise in a circle \\
21 & \texttt{012337} & a person walks forward carefully placing one foot directly in front of the other foot. \\
22 & \texttt{012622} & a person walking in a diagonal line. \\
23 & \texttt{013546} & this person is seated as if playing the drums. \\
24 & \texttt{013735} & a person giving a round of applause \\
25 & \texttt{013855} & person holds left arm out and right arm forward, then shuffles side to side as if doing the chacha dance \\
26 & \texttt{014326} & person lunges forward with left foot first repeatedly \\
27 & \texttt{014541} & a figure does jumping jacks \\
28 & \texttt{M006022} & someone runs backwards in a clockwise motion. \\
29 & \texttt{M013023} & person moves in a clockwise direction in a circle by sprinting \\
30 & \texttt{M013619} & a man swings his left arm back repeatedly. \\
\end{longtable}
\endgroup

\end{document}